\documentclass[conference]{IEEEtran}
\IEEEoverridecommandlockouts
\usepackage{cite}
\usepackage{amsmath,amssymb,amsfonts}
\usepackage{graphicx}
\usepackage{balance}
\usepackage{textcomp}
\usepackage{xcolor}
\def\BibTeX{{\rm B\kern-.05em{\sc i\kern-.025em b}\kern-.08em
    T\kern-.1667em\lower.7ex\hbox{E}\kern-.125emX}}
\begin{document}

\title{Improving Depth Estimation using Location Information}
\author{\IEEEauthorblockN{Ahmed Zaitoon}
\IEEEauthorblockA{\textit{Computer and Sys. Engineering} \\
\textit{Ain-Shams University}\\
Cairo,  Egypt \\
ahmed.zaitoon91@gmail.com}
\and
\IEEEauthorblockN{Hossam El Din Abd El Munim}
\IEEEauthorblockA{\textit{Computer and Sys. Engineering} \\
\textit{Ain-Shams University}\\
Cairo, Egypt \\
hossameldin.hassan@eng.asu.edu.eg}
\and
\IEEEauthorblockN{Hazem Abbas}
\IEEEauthorblockA{\textit{Computer and Sys. Engineering} \\
\textit{Ain-Shams University}\\
Cairo, Egypt \\
hazem.abbas@eng.asu.edu.eg}
}

\maketitle

\begin{abstract}
The ability to accurately estimate depth information is crucial for many autonomous applications to recognize the surrounded environment and predict the depth of important objects. One of the most recently used techniques is monocular depth estimation where the depth map is inferred from a single image.

This paper improves the self-supervised deep learning techniques to perform accurate generalized monocular depth estimation. The main idea is to train the deep model to take into account a sequence of the different frames, each frame is geo-tagged with its location information. This makes the model able to enhance depth estimation given area semantics. We demonstrate the effectiveness of our model to improve depth estimation results. The model is trained in a realistic environment and the results show improvements in the depth map after adding the location data to the model training phase.
\end{abstract}


\IEEEpeerreviewmaketitle

\section{Introduction}

\begin{figure}[!t]
\centerline
{\includegraphics[width=0.5\textwidth]{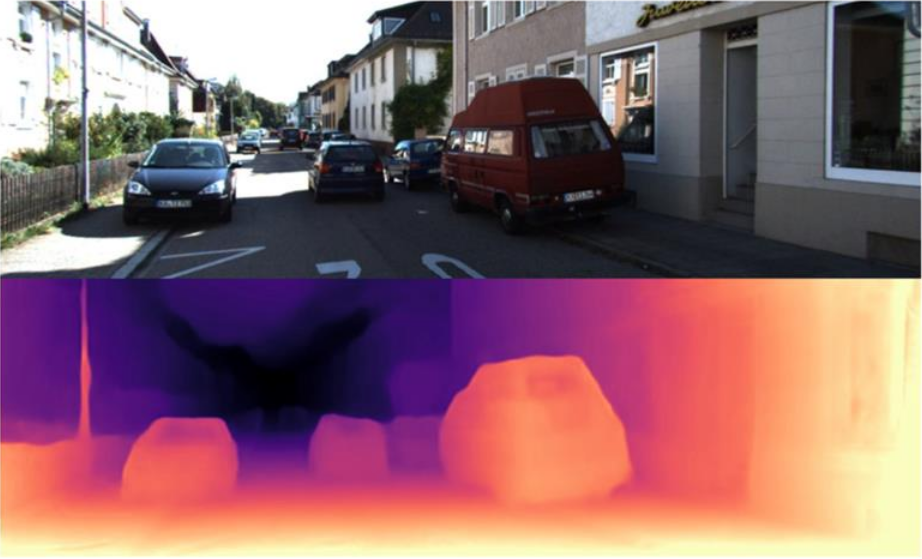}}
\caption{The depth map.}
\label{fig:dm}
\end{figure}

A depth map is an image that depicts the distance between the surfaces of objects from a  given viewpoint, the more the pixel is far the more it will be dark as illustrated in figure \ref{fig:dm}. In computer vision, extracting depth information from images is a fundamental and important task. It is commonly used in navigation \cite{zhu2015vision}, object detection \cite{chai2017obstacle}, semantic segmentation \cite{park2017rdfnet}, and simultaneous localization and mapping (SLAM)  \cite{hu2012robust} among others. In literature, to get depth information from normal plain images, there are three main techniques: geometry-based techniques, sensors-based techniques, and deep learning-based techniques.  

The geometric-based techniques try to recover the three dimensional structure of images from a couple of images based on some geometric constraints. A popular technique in this category is Structure from Motion (SfM) where the images 3D structures are estimated from a series of 2D image sequences \cite{ullman1979interpretation}. It is also applied in SLAM \cite{mur2015orb} and 3D reconstruction \cite{mancini2013using} successfully. Another technique is stereo vision matching, which also can do 3D scene reconstruction using two different views of a scene \cite{zou2010method}\cite{cao2015summary}. These techniques usually depend on obtaining image pairs or image sequences as input %!\cite{mancini2013using}
\cite{cao2015summary}, which makes them need a special setup, opening a question about how to estimate the depth map using only one image. This still is a big challenge.

On the other hand, sensor-based techniques use a special hardware such as RGB-D cameras and LIDAR. RGB-D camera is commonly used in depth estimation due to its ability to obtain depth map on a per-pixel basis of an RGB image directly. However, it suffers from a limited range of measurement and is sensitive to sunlight in the outdoors\cite{tateno2017cnn}, making it suitable for indoor applications only. In outdoor areas, LIDAR is commonly employed in self-driving car industry to measure depth of surrounding objects.  \cite{yoneda2014lidar}. However, it is only able to produce sparse 3D maps, limiting its use in depth estimation field. 

Unlike the previous sensors, monocular camera has gained more attention in depth estimation applications due to its ubiquity, low cost, and small size. So it has been well researched based on deep learning techniques. As with the recent development in deep learning, deep models have shown an impressive performance in many image processing applications, such as image classification \cite{zhang2019fast}, objective detection \cite{pang2019libra}, and semantic segmentation \cite{lyu2019esnet}. A number of neural network models have shown their effectiveness regarding the monocular depth estimation, such as variational auto-encoders (VAEs) \cite{chakravarty2019gen,rizk2019effectiveness}, convolutional neural networks (CNNs) \cite{garg2016unsupervised}, generative adversarial networks (GANs) \cite{aleotti2018generative,boulis2021data}, and recurrent neural networks (RNNs) \cite{wang2019recurrent}.  

In supervised learning techniques\cite{saxena2008make3d}\cite{eigen2014depth}, varied and large  datasets should be collected with correct ground truth which in the case of depth is a formidable challenge. Alternatively, some recent self-supervised methods have shown that monocular depth estimation models can be trained using only monocular videos \cite{zhou2017unsupervised} or synchronized stereo \cite{garg2016unsupervised}\cite{godard2017unsupervised} pairs. Monocular videos techniques try to mitigate the constraints in the self-supervision techniques by using monocular image sequence as it is easy to collect. These techniques try to keep reconstructing the input images from the plain input images and the estimated depth map. Then, we get an accurate depth map when the reconstruction error is small as possible.

Although these self-supervised models solve the need for ground truth depth map, they still need to be improved to provide more accurate depth maps.

This paper introduces a new technique that uses the location information to improve depth estimation results. 
The basic idea is to train the deep model to take into account a sequence of the different frames, each frame is geo-tagged with its location information \cite{shokry2017tale,shokry2020dynamicslam,gu2021effect,shokry2018deeploc}. The location information can improve the depth estimation accuracy as it makes the model able to identify and estimate the common structures, encoded in the location information, on the road images. Based on that, we developed a self-supervised  monocular depth estimation technique. Our technique can easily be integrated with any monocular depth estimation model, which opens the door for more enhancement in depth estimation field.

\section{Related Work}
In this section, we review deep learning models which predicts the per pixel depth map using a single-color image as an input at test time.

\subsection{Supervised Depth Estimation}
It is known that estimating a depth map from only one image is an ill-posed problem as there are multiple depth maps for the same input image.To address this problem, different supervised learning based techniques have shown themselves able to fit predictive models that take advantage of the relationship between coloured images and their depth maps. Examples include various approaches such as combining local predictions \cite{hoiem2005automatic}, \cite{saxena2008make3d}, non-parametric scene sampling \cite{karsch2012depth}, through to end-to-end supervised learning \cite{eigen2014depth}, \cite{laina2016deeper}, \cite{fu2018deep} have been explored in literature. Supervised learning based techniques are also among some of the best performing for stereo estimation \cite{zbontar2016stereo}, \cite{ummenhofer2017demon} and optical flow %!\cite{ilg2017flownet}, 
\cite{wang2018occlusion}.
%! , \cite{mayer2016large} \cite{kendall2017end}

Most of the mentioned techniques are supervised, i.e. they are requiring true depth map in the training phase. However, this is challenging to obtain in reality. i.e. how to obtain the ground truth depth map in the training phase?. To this end, there is an increasing work that exploits weakly supervised training data. This happens in the form of known object sizes \cite{wu2018size}, sparse ordinal depths \cite{zoran2015learning}, \cite{chen2016single}, supervised appearance matching terms \cite{zbontar2016stereo}, \cite{zhan2018unsupervised}, and unpaired synthesized depth data \cite{kundu2018adadepth}, %!\cite{atapour2018real}, \cite{guo2018learning},
 \cite{zou2018df}. All the previous techniques still need to collect an additional depth maps and other annotations. Synthesized training data is a good replacement \cite{mayer2018makes}. However, generating a large amount of synthesized data with a variety of real-world appearance and motion is not easy.

On the other hand, recent work has shown that conventional SfM pipelines can solve this problem by generating a sparse training signal for both depth and camera pose \cite{li2018megadepth},  \cite{klodt2018supervising}, \cite{yang2018deep}, where SfM is typically run as an independent preprocessing stage from learning.

\subsection{Self-supervision Depth Estimation} 
As the true depth map does not exist, another solution is to use self-supervised depth estimation techniques where the training data is either monocular sequences (i.e. a series of images) or stereo image pairs. The model tries to keep reconstructing the input images from the plain input images and the estimated depth map. Then, the accurate depth map is estimated when the reconstruction error is small as possible.

\subsubsection{Stereo pairs-based Training}
The first self-supervised technique takes the pairs of stereo images as input in the offline training phase. Then, by observing the differences between each pair of images, the model is trained to perform monocular depth estimation in the online testing phase. Authors of \cite{xie2016deep3d} presented a discretized depth estimation for the problem of novel view synthesis. Authors of \cite{garg2016unsupervised} further extended this approach by predicting continuous disparity values, and  authors of \cite{godard2017unsupervised} came up with a result that is superior to contemporary supervised techniques by including a left-right depth consistency term. 

Finally, Stereo-based techniques have been further extended with some semi-supervised data \cite{kuznietsov2017semi}, \cite{luo2018single}, additional consistency \cite{poggi2018learning}, generative adversarial networks (GANs) \cite{aleotti2018generative}, \cite{pilzer2018unsupervised},  temporal information %\cite{li2018undeepvo}, \cite{zhan2018unsupervised},
\cite{babu2018undemon}, and, finally,  for real-time use \cite{poggi2018towards}.

\subsubsection{Monocular sequence-based Training}
These techniques try to mitigate the constraints in the self-supervision techniques by using monocular image sequence.   
They use consecutive frame images to train their models. To this end, the model tries to estimate the camera pose 
between frames in the image sequence. This is a challenging problem as objects keep moving.
The camera pose is then used in the training phase to constrain the network.  
In \cite{zhou2017unsupervised} authors proposed the first monocular depth estimation approach where the depth estimation model is trained with
the help of an independent pose estimation network.
Since then, the gap between the stero-based techniques and the monocular-based depth estimation keeps decreasing.

In \cite{yang2017unsupervised}, authors restrict the predicted depth by the predicted surface normals. Same for \cite{yang2018lego} where the edges of the images are used to put a constraint on the predicted depth.

The authors of \cite{mahjourian2018unsupervised} proposed a geometry-based matching technique to enhance depth estimation. 
%
%[43] use a depth normalization layer to overcome the preference for smaller depth values that arises from the commonly used depth smoothness term from [13]. 
%
Finally, authors of \cite{casser2019depth} leverage the predefined masks of commonly exist categories to handle the moving objects problem.

\section{Methodology}

\begin{figure*}[!t]
\centerline
{\includegraphics[width=0.9\textwidth]{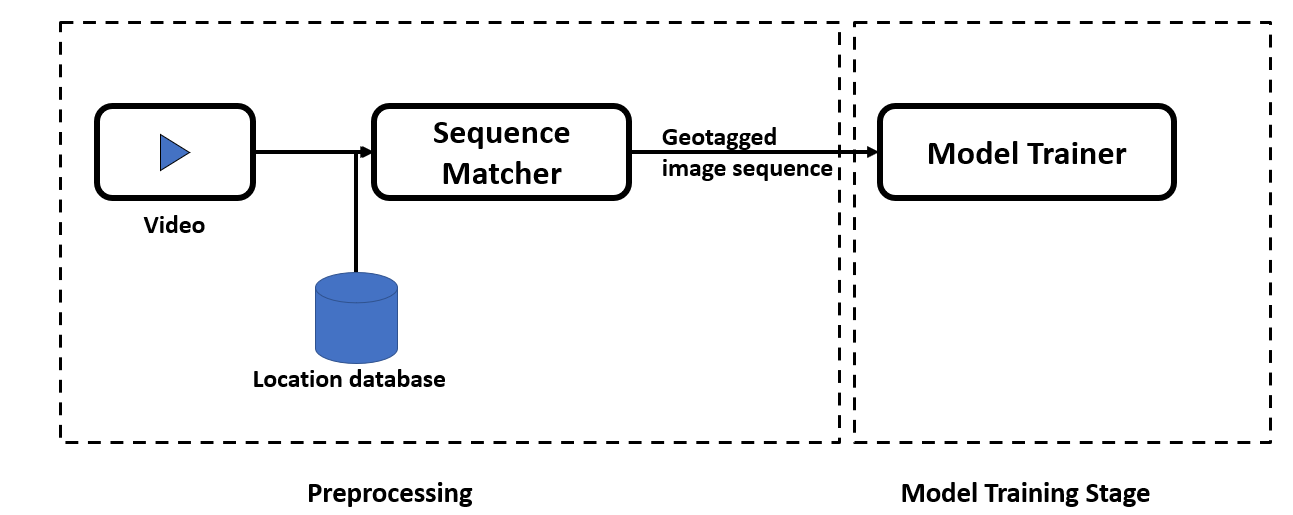}}
\caption{The system architecture}
\label{fig:arch}
\end{figure*}

Figure~\ref{fig:arch} shows our methodology. Our system has two main modules, the preprocessing and the model training module. 

\subsection{Preprocessing}
The input data to the system is a video that has a sequence of images and their location database that contains the latitude and longitude for each frame. First, we sample the video to a sequence of images with suitable time difference between frames. Then, the sequence matcher algorithm links the location information (latitude and longitude) to its frame using frames timestamps.
Finally, the sequence of geotagged images is fed to the model training phase. 

\subsection{Model training}
\label{mdl}
In order to embed the location information to the input of the deep model, we encode it in the images Alpha Channel without affecting the original features of the images. The geotagged images then become the input of the depth estimation model.

%In order to embed the location information to the input of the deep model, we used a single feed-forward fully connected layer. The input to the fully connected layer is the image pixels and the location information. The output has the same size as the image dimensions (number of row pixels times number of column pixels). The output of the layer is then feed to the depth estimation model.
%

We use one of the  most robust networks for self-supervised depth estimation \cite{godard2019digging}. Our deep model is self-supervision which uses monocular geotagged sequence of images to provide the training data. Here, the network must not only predict the depth map, but also the camera pose between frames, which is difficult in the presence of moving objects. the estimated pose of the camera is only used to constrain the network during training. As the image sequence has geographical information, the deep model is trained better to estimate the road features depth map. 

The used depth estimation network is based on the general U-Net architecture which is an encoder-decoder network with skip connections that allows us to represent local information as well as deep abstract features. Our encoder is a ResNet18, which has 11M parameters. We start with ImageNet pretrained weights. The depth decoder is similar to \cite{godard2017unsupervised}, with sigmoids at the output and ELU nonlinearities elsewhere.

\begin{figure*}[!t]
\centerline
{\includegraphics[width=1\textwidth]{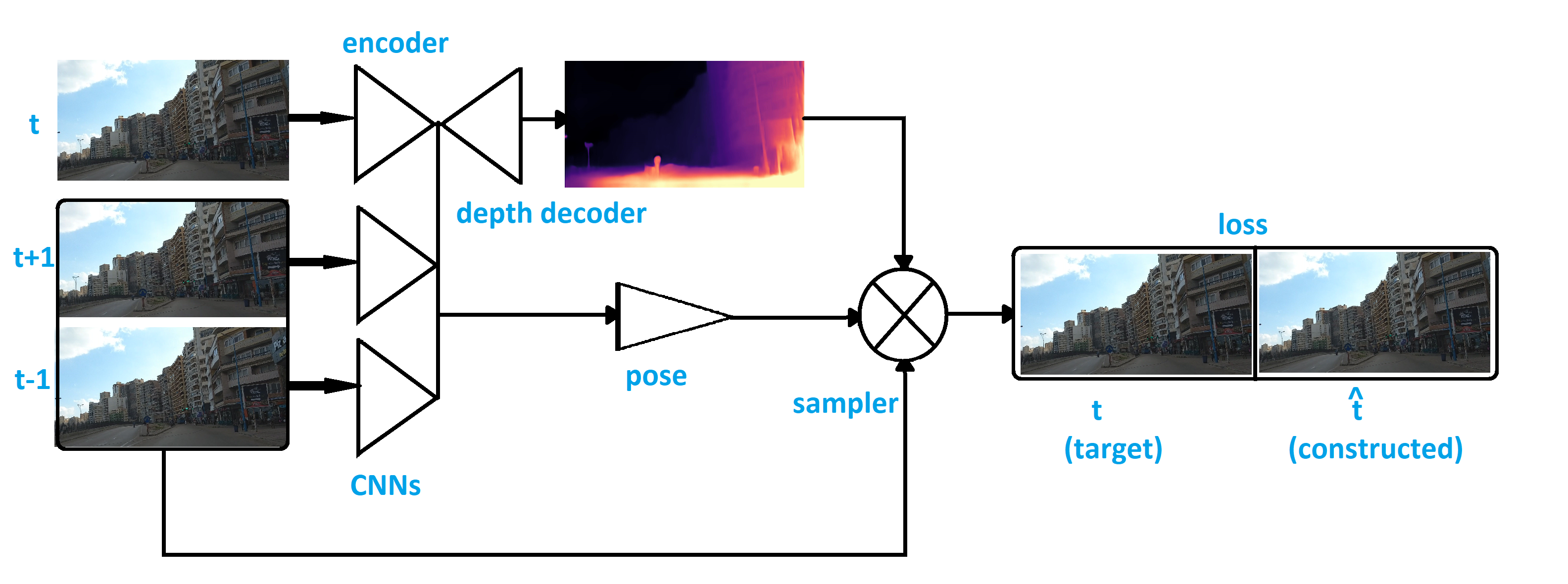}}
\caption{The network architecture}
\label{fig:net}
\end{figure*}

Figure \ref{fig:net} shows our deep model in the training phase. The model takes three frames at time $t$, $t+1$, and $t-1$. The frame at time $t$ is fed to an autoencoder network for depth estimation while the remaining frames, at time $t-1$ and $t+1$, are fed to the encoder network for pose estimation. The basic idea is to use the frames at times $t-1$ and $t+1$ to reconstruct the frame at time $t$. Then, by comparing the reconstructed image at time $t$ and the original image at time $t$, the network calculates the reconstruction loss and keeps modifying the estimated depth map to minimize image reconstruction loss. At this stage, the network can benefit from the location information to update the depth map by putting a constraint on reconstruction loss. 
 
Mathematically speaking, our self-supervised method is trained by monocular image sequences, and the projection between adjacent frames is used to build the geometric constraints as follows:

\begin{equation}
p_{n-1} \sim K T_{n\rightarrow n-1} D_n(p_n) K^{-1} p_n
\end{equation} 

where $p_n$ refers to the pixel on image In, and $p_{n-1}$ stands for the corresponding pixel of $p_n$ on image $I_{n-1}$. $K$ is intrinsics matrix for the camera, which is known in advance. $D_n(p_n)$ signifies the value of the depth at pixel $p_n$, and $T_{n\rightarrow n-1}$ refers to the spatial transformation between $I_n$ and $I_{n-1}$. Therefore, if we know $T_{n\rightarrow n-1}$ and $D_n(p_n)$, the correspondence between the pixels on the two images ($I_n$ and $I_{n-1}$) can be established using projection function.

The depth network is designed to take a single image and predict the depth map $\hat{D_n}$, and regress the transformation $T_{n\rightarrow n-1}$ between images ($I_n$ and $I_{n-1}$) using a pose network.
The pixel correspondences between frames $I_n$ and $I_{n-1}$ are built up Based on the output of depth and pose networks:

\begin{figure*}[!t]
	\centerline
	{\includegraphics[width=0.9\textwidth]{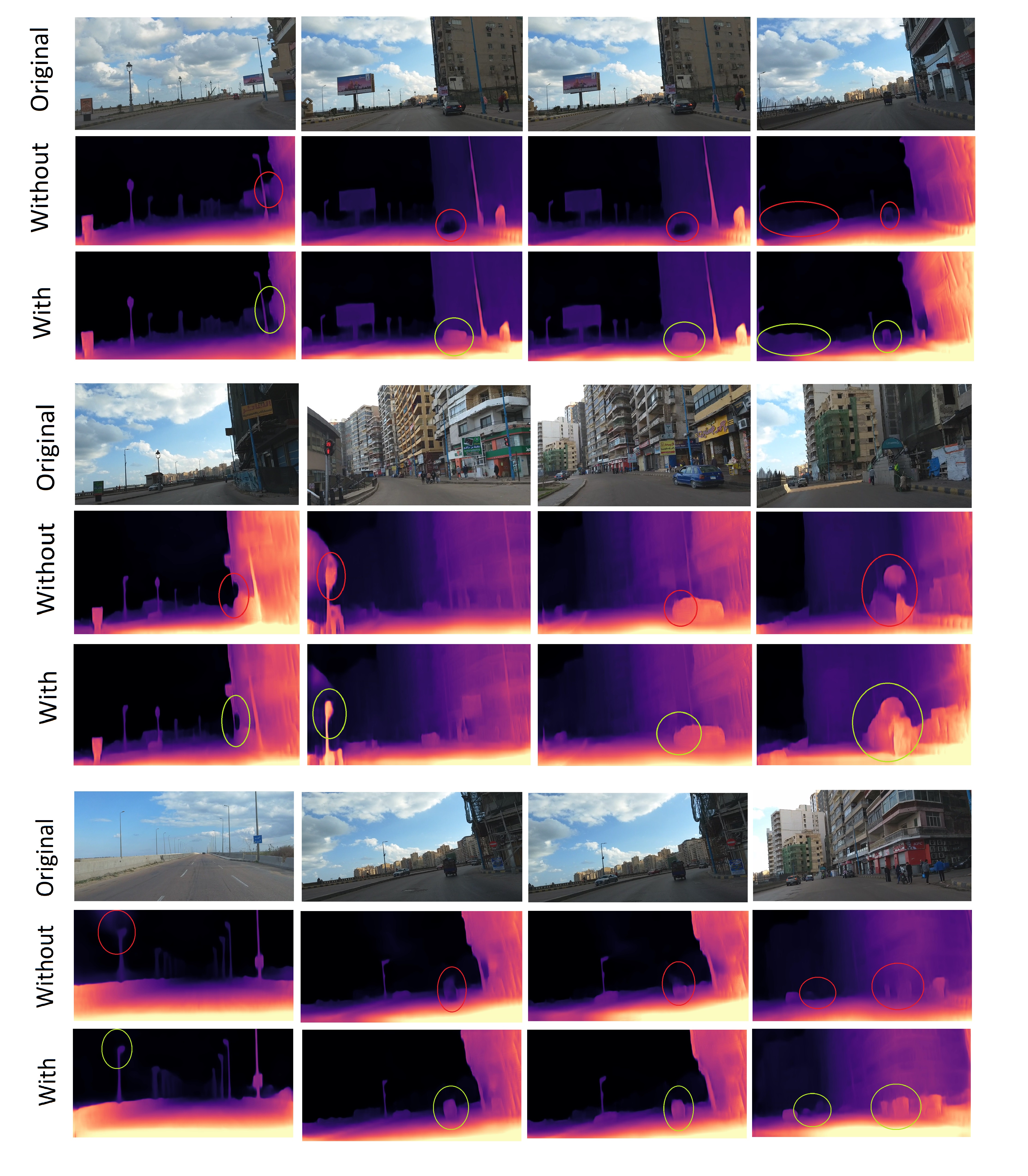}}
	\caption{The depth estimation map for some test samples after training the network once by using a geotagged image sequence and once using an image sequence without the location information.}
	\label{res}
\end{figure*}

\begin{equation}
p_{n-1} \sim K \hat{T}_{n\rightarrow n-1} \hat{D}_n(p_n) K^{-1} p_n
\end{equation} 

As the geometric constraints ,the photometric error between the corresponding pixels is calculated as in \cite{godard2019digging}. Based on the projection function, the frame $\hat{I}_n(p)$ is reconstructed from $I_{n-1}$  using a view reconstruction algorithm . Although during training, the depth network and pose network is used together, they can be used separately during testing.  

To summarise, depth network and pose network predict the depth $D_n$ and the pose $\hat{T}_{n\rightarrow n-1}$, which are used to determine the projection relationship between images $I_n$ and $I_{n-1}$, and hence, based on the projection, the image warping process is used to reconstruct $\hat{I}_n$. by calculating the differences between real $I_n$ and reconstructed $\hat{I}_n$ images , we can perform self-supervised training to the network.
%\vspace{2cm}

\section{Experimental Evaluation}
To validate our model, we collected four independent datasets. All datasets are collected with the location information. Two of them at a sea road and two at a highway road. We train the model described in section \ref{mdl} using samples from sea and highway roads in two datasets and test it using samples from the other two (sea and highway roads datasets). We repeat the experiment twice,  once using geotagged images and another one using images without the location information.

Since there is no ground truth depth map dataset with location information, we follow the convention from literature that presents output samples for depth maps to show the improvement of the depth estimation technique. Figure \ref{res} shows the depth estimation map for some test samples using the geotagged image sequence (labeled \textit{with}) and once using image sequence without the location information (labeled \textit{without}). The red circles show a limitation in the model that does not take the spatial information while the green circles show how our model can solve these limitations. The location information not only can improve the depth map for the road semantics, but also the depth map for the dynamic objects (for example the last image in the first row). The figure shows that location information can significantly improve the depth estimation results. 

\section{Conclusion}
In this paper, we introduced a new technique to improve monocular depth estimation by developing a deep model that takes into account the spatial information. The basic idea is to train the deep model to take into account a sequence of the different frames, each frame is geo-tagged with its location information. This makes the model able to enhance the depth estimation given the geographical area semantics. We showed the effectiveness of our model in a realistic environment and the results show improvements in the depth maps after adding the location data to the model training phase. Our technique can be used with any monocular depth estimation model, which opens the door for more enhancement in depth estimation field.

\balance
\bibliographystyle{IEEEtran}
\bibliography{zaitoon_icc}

\end{document}